\documentclass{article}

\usepackage{PRIMEarxiv}
\usepackage[utf8]{inputenc}
\usepackage[T1]{fontenc}
\usepackage{graphicx}
\usepackage{mathptmx}
\usepackage{times}
\usepackage{amsmath,amssymb,amsfonts}
\DeclareMathAlphabet{\mathcal}{OMS}{cmsy}{m}{n}
\usepackage{bm}
\usepackage{array,booktabs,multirow}
\usepackage{float}
\usepackage{adjustbox}
\usepackage{algorithm}
\usepackage{algpseudocode}
\usepackage{nicefrac}
\usepackage{microtype}
\usepackage{fancyhdr}
\usepackage{url}
\usepackage{hyperref}
\usepackage{cleveref}

\algrenewcommand\algorithmicrequire{\textbf{Input:}}
\algrenewcommand\algorithmicensure{\textbf{Output:}}
\newcommand{\task}{r^{\mathrm{task}}}
\newcommand{\shape}{r^{\mathrm{shape}}}
\newcommand{\eps}{\varepsilon}

\title{Large Language Model Guided Incentive Aware Reward Design for Cooperative Multi-Agent Reinforcement Learning}

\author{
  Dogan Urgun \\
  Department of Electrical and Electronics Engineering \\
  Karabuk University \\
  78050 Karabuk, T\"urkiye \\
  \texttt{durgun@karabuk.edu.tr} \\
  \And
  Gokhan Gungor\\
  Department of Mechatronics Engineering \\
  Karabuk University \\
  78050 Karabuk, T\"urkiye \\
  \texttt{gokhangungor@karabuk.edu.tr} \\
}

\begin{document}
\maketitle
\thispagestyle{empty}

\begin{abstract}
In cooperative multi-agent reinforcement learning (MARL), intermediate rewards are commonly added to a sparse task reward to guide learning. Such rewards, however, also shape the incentives of individual agents, since they may reward events that cannot contribute to task success or assign the credit for a joint outcome to only one of the contributing agents. Large language models (LLMs) can generate reward functions as executable code, but they typically refine them from summary statistics of training, which do not reveal such misaligned incentives. To address this limitation, this paper proposes incentive-aware reward repair. A task-specific diagnostic module, fixed before the search, measures which actions a candidate reward favors and how it distributes credit among agents, and an LLM uses these measurements to revise the terms of an executable reward program. All candidates are trained with a common multi-agent proximal policy optimization (MAPPO) learner that retains agent-specific rewards, and the final reward is selected by task return alone. In two Overcooked layouts and in Level-Based Foraging, the repaired rewards achieved the highest task return among all compared designs, including an otherwise identical LLM agent without the module. Moreover, the same manual shaping prevented cooperation when it credited only the acting agent and enabled it when it was shared, and reassigning placement credit within the repaired reward almost eliminated cooperation, which shows that the allocation of intermediate credit, not only its amount, can determine whether cooperation is learned.
\end{abstract}

\keywords{Artificial intelligence \and cooperative systems \and credit assignment \and large language models \and multi-agent reinforcement learning \and reward shaping}

\section{Introduction}

Cooperative multi-agent reinforcement learning (MARL) is used to train teams of agents that act jointly toward a shared objective~\cite{gronauer2022survey}. The reward function specifies this objective and, implicitly, how credit for a joint outcome is distributed among the individual agents and their decisions~\cite{wong2023deep}. When only task completion is rewarded, this distribution is difficult to learn. Success follows from a long sequence of interdependent actions by several agents, and it is rarely observed early in training~\cite{sutton2018reinforcement,foerster2018counterfactual}.

Reward shaping addresses this difficulty by adding intermediate rewards for events that are expected to precede success~\cite{ng1999shaping}. Under specific conditions, potential-based shaping preserves the optimal policies of the original task and, in multi-agent systems, its Nash equilibria~\cite{ng1999shaping,devlin2011theory}. Most practical shaping, however, consists of event bonuses that carry no such guarantee, and a bonus that is effective in one context can induce unintended behavior in another~\cite{hadfieldmenell2017ird}. Cooperative tasks raise two further questions. The first is whether a rewarded event actually advances the shared task, and the second is which agent receives the credit for it. If an intermediate reward favors an unproductive action, or credits only one of the agents whose actions produced an outcome, the incentives of the individual agents diverge from the objective of the team~\cite{wolpert2002coin,devlin2014pbdR}.

Both forms of divergence arise in Overcooked, a widely used benchmark for cooperative coordination~\cite{carroll2019overcooked}. In version 1.1.0 of its simulator~\cite{overcookedAI}, cooking can start before a recipe is complete, and the built-in shaping rewards the pickup of a cooked soup whether or not the soup matches the current order. A cook can therefore be rewarded for a soup that has no delivery value. In the Forced Coordination layout, ingredients must be passed across a counter, yet the placement bonus is given only to the agent that puts an ingredient into the pot. When each agent learns from its own reward~\cite{tan1993independent}, the agent that supplies the ingredients therefore receives no credit for its contribution. Neither problem concerns the magnitude of the reward. Both concern the incentives that it creates.

Large language models (LLMs) have recently been used to automate reward design by writing reward functions as executable code. Dense reward code has been generated from environment abstractions~\cite{xie2024text2reward}, refined through reflection on reward-component statistics collected during training~\cite{ma2024eureka}, and improved through structured search over candidate designs~\cite{li2025rstar,gao2025rfagent}. LLM-generated rewards have also been applied to cooperative MARL, for example in multi-robot cooperation~\cite{zhu2025lamarl}, credit assignment~\cite{qu2025latent}, and platoon coordination~\cite{wei2026platoon}. The feedback used to refine these rewards typically consists of summary statistics, such as task returns and the magnitudes of reward components~\cite{ma2024eureka,wei2026platoon}. Such statistics show whether a reward is effective, but not which rewarded actions conflict with the task or which agent receives the credit for a cooperative event. Agentic approaches that allow the LLM to inspect source code and evaluation records broaden this evidence~\cite{li2026arbo}, but they leave the detection of misaligned incentives to the model itself. In addition, many cooperative learners optimize a single team reward~\cite{sunehag2018value,yu2022surprising}. Rewards that differ only in how credit is allocated among the agents then produce the same learning signal, and the allocation chosen by the designer has no effect on what is learned.

\begin{figure}[!htbp]
\centering
\includegraphics[width=\linewidth,height=0.70\textheight,keepaspectratio]{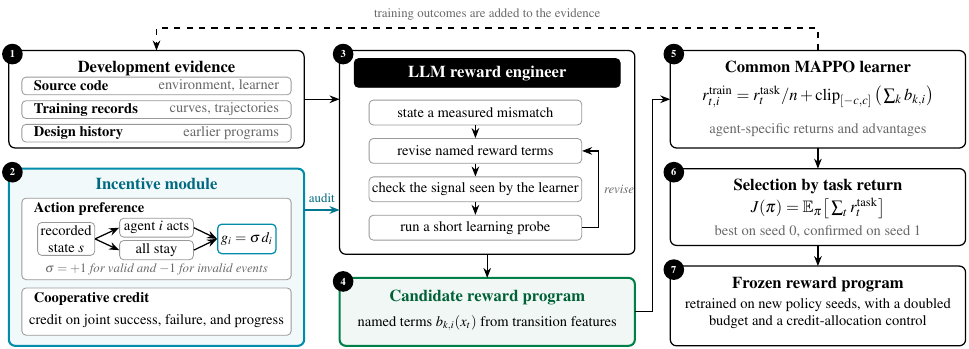}
\caption{Overview of incentive-aware reward repair. An LLM reward engineer (3) revises candidate rewards (4) using development evidence (1) and a fixed incentive module (2), which the ablation agent does not receive. Candidates are trained with a common learner (5) and selected by task return (6), and the selected program is frozen and retrained (7).\label{fig:overview}}
\end{figure}

To address these limitations, this paper proposes incentive-aware reward repair, illustrated in Fig.~\ref{fig:overview}. For each task, a fixed diagnostic module is defined before the reward search. It measures whether a candidate reward prefers task-relevant actions over unproductive ones in recorded states, and how it distributes credit over coordination events. An LLM reward engineer uses these measurements to locate misaligned terms in an executable reward program, revises them, and tests each revision through short learning probes. All candidates are trained with the same multi-agent proximal policy optimization (MAPPO) learner~\cite{yu2022surprising}, which retains the reward of each agent in its own return and advantage, and the final reward is selected by task return alone. The method is evaluated on two Overcooked layouts, Forced Coordination and Asymmetric Advantages~\cite{carroll2019overcooked}, and on Level-Based Foraging (LBF)~\cite{christianos2020shared,papoudakis2021benchmarking}, three tasks that reward coordinated behavior.

The main contributions of this paper are as follows.
\begin{enumerate}
\item Incentive-aware reward repair is introduced, in which fixed task-specific measurements of action preference and cooperative credit guide an LLM in revising executable reward programs, while the learner and the selection rule are held fixed.
\item A controlled evaluation on three cooperative tasks is conducted, in which the contribution of the diagnostic module is separated from that of the LLM agent, and the selected rewards are retrained on new seeds and under a doubled training budget.
\item The allocation of intermediate credit among agents is identified as a distinct dimension of cooperative reward design. For learners that retain agent-specific rewards, this allocation can determine whether cooperation is learned even when the total reward of the team is unchanged.
\end{enumerate}

To support reproducibility, the implementation and the selected reward programs are publicly available at \href{https://github.com/DoganUrgun/incentive-aware-reward-design}{\nolinkurl{github.com/DoganUrgun/incentive-aware-reward-design}}. The remainder of this paper is organized as follows. Section~\ref{sec:related} reviews related work, and Section~\ref{sec:method} presents the methodology. Section~\ref{sec:setup} describes the experimental setup, and Section~\ref{sec:results} reports the results. Section~\ref{sec:discussion} discusses the findings and their limitations, and Section~\ref{sec:conclusion} concludes the paper.

\section{Related Work}
\label{sec:related}
This work builds on three lines of research. Reward shaping provides intermediate learning signals, credit assignment distributes a joint outcome among cooperating agents, and LLM-based reward design automates the construction of reward functions. The following subsections review each line and identify the aspect that the present work addresses.

\subsection{Reward Shaping and Misaligned Incentives}
Reward shaping supplements a sparse task reward with intermediate signals, but the added signal can also change which behavior is optimal. Ng et al. showed that shaping derived from a potential function preserves the optimal policies of the original task~\cite{ng1999shaping}. This result was later extended to the Nash equilibria of multi-agent systems~\cite{devlin2011theory} and to arbitrary reward functions expressed as potential-based advice~\cite{harutyunyan2015pba}. These guarantees, however, require the shaping to take a potential-based form, whereas practical shaping often consists of event bonuses, such as the built-in rewards of Overcooked~\cite{carroll2019overcooked}. When such bonuses are misspecified, agents can satisfy them in ways the designer did not intend~\cite{hadfieldmenell2017ird}. Rather than restricting the form of the reward, the present work measures the incentives that a candidate reward creates and uses these measurements to repair it.

\subsection{Credit Assignment in Cooperative MARL}
In cooperative MARL, the success of the team must be attributed to the decisions of individual agents. Most methods address this problem within the learning algorithm while optimizing a shared team reward. COMA estimates the contribution of each action with a counterfactual baseline~\cite{foerster2018counterfactual}, value-factorization methods decompose a joint value function into per-agent components~\cite{sunehag2018value,rashid2020qmix}, and MAPPO learns from the team reward with a centralized critic~\cite{yu2022surprising}. Difference rewards take a complementary approach and give each agent a reward that reflects its marginal contribution to the team~\cite{wolpert2002coin,devlin2014pbdR}, and LLMs have also been used to derive latent rewards for credit assignment~\cite{qu2025latent}. In these methods, credit is assigned by the learner or by a dedicated signal, and the credit distributed by the shaping reward itself has received less attention. In this work, the learner is held fixed and retains the reward of each agent, as in independent learning~\cite{tan1993independent}, so the credit allocation encoded in a reward program becomes a property that can be measured and revised.

\subsection{Reward Design With Large Language Models}
LLMs can translate task descriptions and environment code into executable reward functions. Text2Reward generated dense reward code from environment abstractions and refined it with human feedback~\cite{xie2024text2reward}. Eureka introduced an iterative scheme in which reward programs are generated, evaluated through training, and revised by reflection on reward-component statistics~\cite{ma2024eureka}, and an adaptation of this scheme serves as a reference condition in the present experiments. Later studies refined this search through reward-structure evolution~\cite{li2025rstar}, tree search~\cite{gao2025rfagent}, and agentic optimization with source inspection~\cite{li2026arbo}. In cooperative settings, LLMs have generated rewards and prior policies for multi-robot cooperation~\cite{zhu2025lamarl} and rewards for platoon coordination~\cite{wei2026platoon}. Across these studies, the LLM receives evidence about training outcomes and must infer by itself which incentives cause a failure. The present work instead provides explicit measurements of action preference and cooperative credit, and assesses their value against an LLM agent that has the same tools but not these measurements.

\section{Methodology}
\label{sec:method}
This section describes the four components of the method, namely the reward contract, the incentive module, the repair procedure, and the policy learner. Their interaction is illustrated in Fig.~\ref{fig:overview}.

\subsection{Reward Contract}
All reward conditions share a common task objective and shaping
interface. Let $s_t$ denote the joint state, $o_{t,i}$ the
observation of agent $i$, and $\mathbf a_t$ the joint action
of $n$ agents. Performance is measured by the native,
undiscounted task return
\begin{equation}
J(\pi)=\mathbb E_{\tau\sim\pi}\left[
\sum_{t=0}^{T-1}\task(s_t,\mathbf a_t,s_{t+1})
\right],
\end{equation}
where $\pi$ denotes the joint policy, $\tau$ an episode
trajectory, and $T$ the episode horizon. The task reward
at step $t$ is $\task_t=\task(s_t,\mathbf a_t,s_{t+1})$.
This return is used for selection and evaluation regardless
of the shaping applied during training.

A reward program returns between 1 and 12 named additive
terms $b_{k,i}(x_t)$ computed from numeric transition features
$x_t$. The shaping and training rewards of agent $i$
are defined, respectively, as
\begin{equation}
\begin{aligned}
\shape_{t,i}&=\operatorname{clip}_{[-c,c]}
\left[\sum_k b_{k,i}(x_t)\right],\\
r^{\mathrm{train}}_{t,i}&=\task_t/n+\shape_{t,i}.
\end{aligned}
\label{eq:training_reward}
\end{equation}
The task reward therefore enters the team sum exactly once.
The clipping bound is $c=8$ in Overcooked and $c=1$ in
Foraging. A logged clipping adjustment ensures that the
reported terms sum to the shaping reward used for training.

The features describe positions, held objects, pot and food states, recipe validity, distances, and partner targets before and after each transition, without changing observations or dynamics. Programs are written in a restricted NumPy language and must pass checks of schema, determinism, finiteness, and magnitude.

\subsection{Incentive Module}
The incentive module examines the incentives that a candidate reward creates. In Overcooked, it measures which recipe actions the reward prefers locally, and in Foraging, it measures how the reward distributes credit across coordination events. Both diagnostics are computed on states recorded during development and are fixed before the search.

Action preference is measured by replaying a recorded state $s$ in the unmodified simulator under two joint actions, yielding single-step transitions $\xi^{\mathrm{int}}$ and $\xi^{\mathrm{stay}}$. In $\xi^{\mathrm{int}}$, agent $i$ interacts while its partner stays, and in $\xi^{\mathrm{stay}}$, both agents stay. Each case receives a reference sign $\sigma(s,i)$ from the task event that the interaction produces, and the reward difference of the acting agent and its signed gap are
\begin{align}
d_i(p,s)&=r^{\mathrm{train}}_{i}(\xi^{\mathrm{int}};p)
          -r^{\mathrm{train}}_{i}(\xi^{\mathrm{stay}};p),\\
g_i(p,s)&=\sigma(s,i)\,d_i(p,s)
\end{align}
for a candidate program $p$, where $r_i^{\mathrm{train}}(\xi;p)$ denotes the training reward of agent $i$ for transition $\xi$ under program $p$. In Overcooked, $\sigma=+1$ for starting or collecting an exact positive-value recipe, and $\sigma=-1$ for starting an incomplete or invalid recipe or collecting a soup without order value. A gap above $\eps=10^{-8}$ is aligned with the reference, a gap below $-\eps$ is opposed, and the remaining gaps are indifferent. An opposed or indifferent gap marks a term to investigate, and any repair is still tested through learning, since the sign expresses a local preference rather than the optimal long-horizon action. Cooperative credit is measured by grouping recorded transitions according to their contribution to joint success. In Foraging, two level-1 agents must load a level-2 food together. Let $\mathcal F(s)$ denote the set of active foods and $D_i(f,s)$ the Manhattan distance
from agent $i$ to food $f$. The common-target distance is
\begin{equation}
D^{\mathrm{joint}}(s)=\min_{f\in\mathcal F(s)}\max_i D_i(f,s)
\end{equation}
This is the distance of the farther agent to the food
that both agents can reach soonest. The module reports the shaping assigned to successful joint loads, unsuccessful loads, explicit idling, progress in $D^{\mathrm{joint}}$, and progress toward an agent's own nearest food that increases $D^{\mathrm{joint}}$. These observational strata indicate where a reward should be inspected rather than prescribe a penalty.

\subsection{Reward Repair Procedure}
Reward programs are developed iteratively from an initial reward, as summarized in Algorithm~\ref{alg:repair}. The search runs for $G$ rounds with $K$ candidates per round. For each candidate, the LLM reward engineer examines the module output and the development records of the current parent, states a measured mismatch and its expected effect on behavior, and revises the responsible terms. Before training, a transition check reports which terms activate and how the signal of each agent changes, and the revision can then be tested in short learning probes.

Each completed candidate is trained with the full budget on a development seed, and the program with the highest task return becomes the parent of the next round. After the last round, the best program is retrained on a second seed and replaces the initial reward only if its two-seed mean is strictly higher. Here, $J_0(q)$ denotes the task return of reward program $q$ on development seed 0, $\bar J(q)$ its mean task return over development seeds 0 and 1, and $\mathcal A$ the development archive indexed by reward programs.

\begin{algorithm}[t]
\caption{Incentive-Aware Reward Repair}
\label{alg:repair}
\small
\begin{algorithmic}[1]
\Require initial reward $p_0$, incentive module $\mathcal M$, rounds $G$, candidates per round $K$
\Ensure frozen reward program $p^\star$
\State $p \gets p_0$, \ $\mathcal A \gets$ development runs of $p_0$
\For{$g = 1, \dots, G$}
  \For{$k = 1, \dots, K$}
    \State Evaluate $p$ with $\mathcal M$ and inspect $\mathcal A$
    \State Revise the mismatched terms of $p$ into $p_{g,k}$
    \State Validate $p_{g,k}$ and check its training signal
    \While{revision is needed and probes remain}
      \State Probe $p_{g,k}$ briefly and revise it
    \EndWhile
    \State Train $p_{g,k}$ on seed 0 and add the run to $\mathcal A$
  \EndFor
  \State $p \gets \arg\max_{q\in\mathcal A} J_0(q)$ \Comment{seed-0 return}
\EndFor
\State Train $p$ on seed 1
\State $p^\star \gets \arg\max_{q\in\{p,\,p_0\}} \bar J(q)$ \Comment{two-seed mean}
\State \Return $p^\star$
\end{algorithmic}
\end{algorithm}

\subsection{Policy Learner}
All candidates are trained with the same learner, so that differences in outcome can be attributed to the reward. The learner is MAPPO~\cite{yu2022surprising} with a shared actor $\pi_\theta(a_i\mid o_i)$ and a centralized critic $V_{\phi,i}(z_t)$ that has one output per agent, where $z_t$ concatenates the agent observations. Each agent retains its own training reward from~\eqref{eq:training_reward}, and generalized advantage estimation is computed per agent,
\begin{align}
\delta_{t,i}&=r^{\mathrm{train}}_{t,i}
  +\gamma m_tV_{\phi,i}(z_{t+1})-V_{\phi,i}(z_t),\\
\widehat A_{t,i}&=\delta_{t,i}
  +\gamma\lambda m_t\widehat A_{t+1,i},
\end{align}
where $m_t$ is zero at the end of an episode and one otherwise. The actor update of each agent uses its own advantage, the critic output of each agent regresses its own return, and advantages are normalized jointly over all agents.

Retaining agent-specific rewards makes credit allocation visible to learning, since moving shaping between agents changes their individual advantages even when the team total is unchanged. Each training run uses 200 updates with eight parallel environments and 256 steps per environment, so one full budget is $B=409{,}600$ transitions. Actor and critic are two-layer networks with 128 tanh units, trained with Adam at a learning rate of $3\times10^{-4}$ for four PPO epochs per update, with minibatches of 256, clipping ratio 0.2, discount 0.99, GAE parameter 0.95, entropy coefficient 0.01, and gradient norm bound 0.5.

\section{Experimental Setup}
\label{sec:setup}
The experiments evaluate the proposed method against alternative reward designs under a common learner and training budget. This section describes the tasks, compared conditions, search protocol, and metrics, followed by two additional experiments.

\begin{table}[!htbp]
\centering
\caption{Reward Conditions in the Primary Comparison\label{tab:conditions}}
\small
\setlength{\tabcolsep}{4pt}
\renewcommand{\arraystretch}{1.15}
\begin{adjustbox}{max width=\linewidth}

\begin{tabular}{@{}lcccc@{}}
\toprule
Condition & LLM & \shortstack{Agent\\tools} & \shortstack{Incentive\\module} & \shortstack{Search\\budget}\\
\midrule
Task only & -- & -- & -- & --\\
Manual & -- & -- & -- & --\\
Eureka pattern & \checkmark & -- & -- & $6B$\\
Agent without module & \checkmark & \checkmark & -- & $6B$\\
Agent with module & \checkmark & \checkmark & \checkmark & $6B$\\
\bottomrule
\end{tabular}
\end{adjustbox}

\vspace{4pt}
\begin{minipage}{\linewidth}\footnotesize
All LLM conditions use the same model, reward interface, and learner. Agent tools comprise source and record access, numerical analysis, and learning probes. $B$ denotes one training run of 409,600 transitions.
\end{minipage}
\end{table}

\subsection{Tasks}

\begin{figure}[!htbp]
\centering
\begin{minipage}[t]{0.32\linewidth}
\centering
\includegraphics[width=\linewidth,height=2.55cm,keepaspectratio]{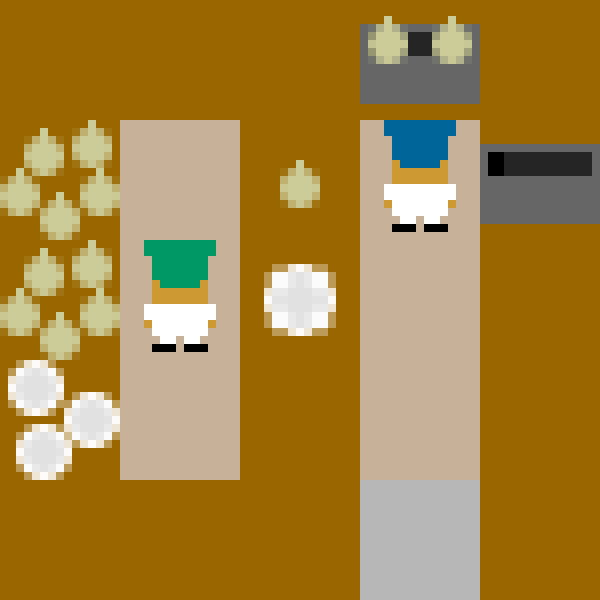}
\par\scriptsize (a) Forced Coordination
\end{minipage}\hfill
\begin{minipage}[t]{0.32\linewidth}
\centering
\includegraphics[width=\linewidth,height=2.55cm,keepaspectratio]{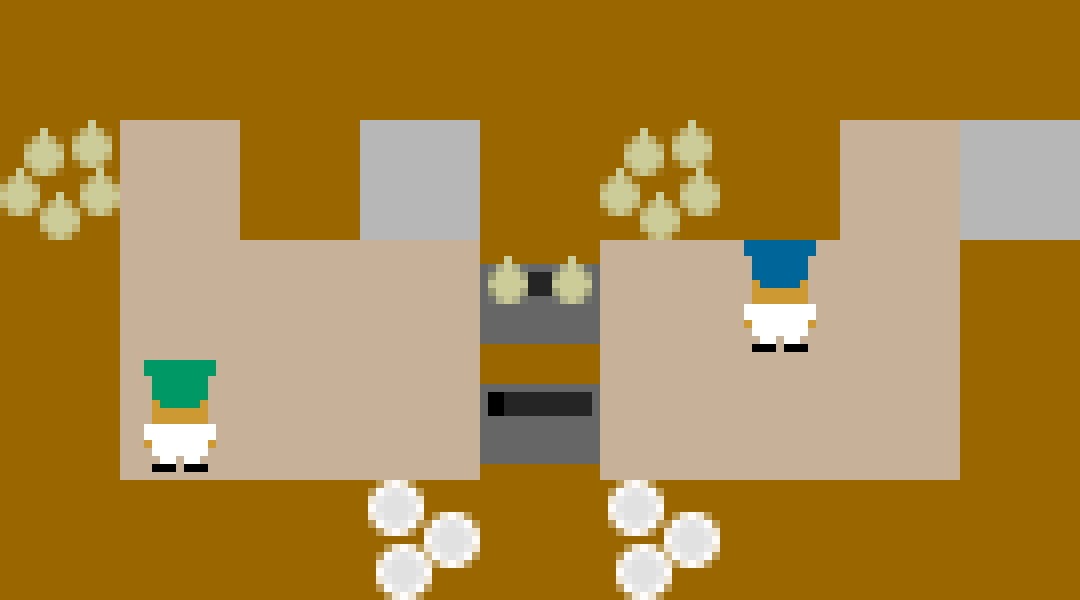}
\par\scriptsize (b) Asymmetric Advantages
\end{minipage}\hfill
\begin{minipage}[t]{0.32\linewidth}
\centering
\includegraphics[width=\linewidth,height=2.55cm,keepaspectratio]{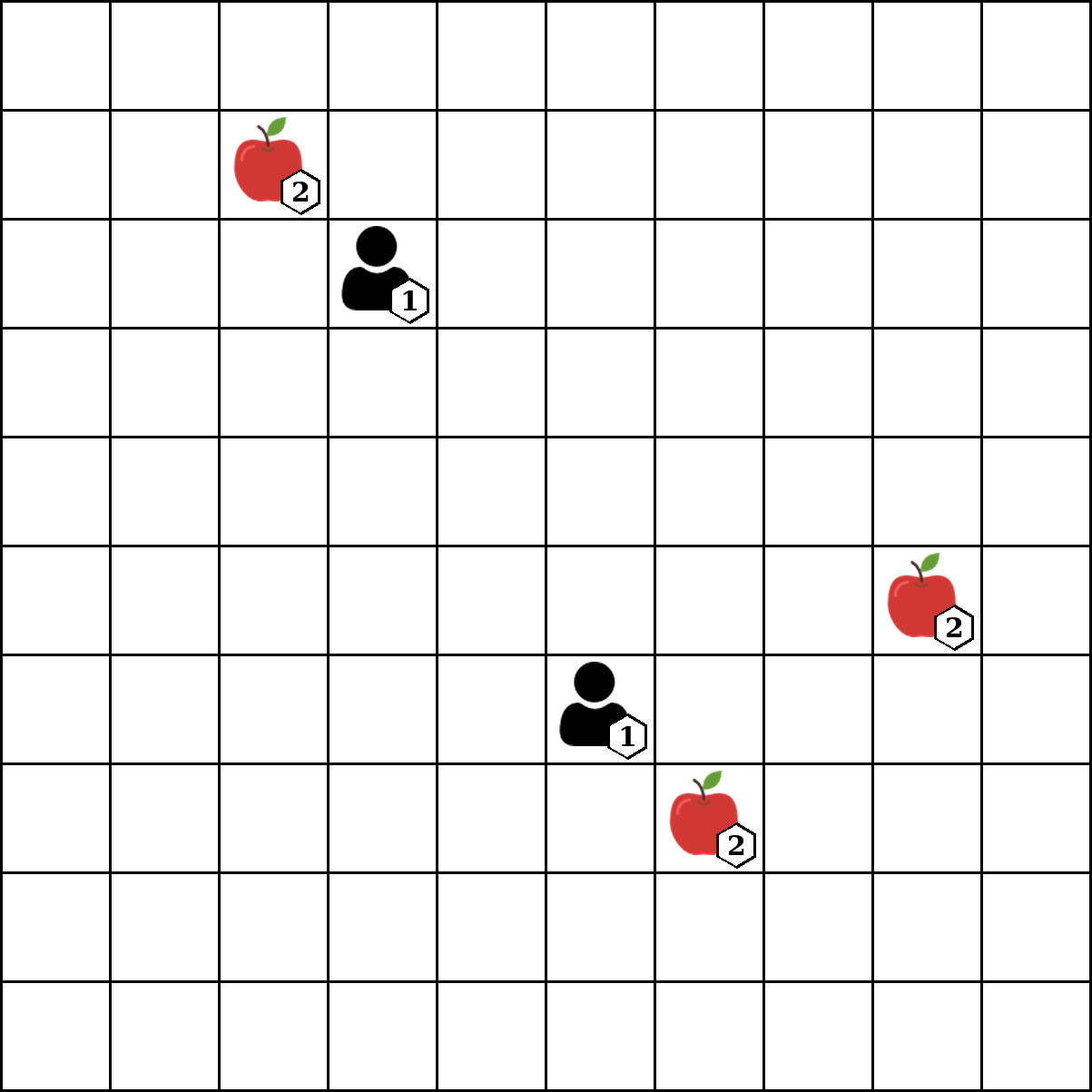}
\par\scriptsize (c) Level-Based Foraging
\end{minipage}
\caption{Benchmark tasks used in the experiments.\label{fig:tasks}}
\end{figure}

Three cooperative tasks were used, comprising two Overcooked layouts with long episodes and delayed rewards and a foraging task with short episodes (Fig.~\ref{fig:tasks}). The first is the Forced Coordination layout of Overcooked-AI 1.1.0~\cite{carroll2019overcooked,overcookedAI}, in which a counter separates the two chefs, so that ingredients and dishes must be passed from one side to the other. Episodes last 400 steps, and only a three-onion soup earns the delivery reward of 20, although cooking can start before a recipe is complete. The native shaping gives 3 for placing an ingredient, 3 for a useful dish pickup, and 5 for picking up a cooked soup, each credited to the acting chef. The manual reward uses the same coefficients but divides this shaping equally between the two chefs. In contrast to the upstream training wrapper, which gives each chef the full team reward~\cite{overcookedRewardWrapper}, the task reward is split equally as in~\eqref{eq:training_reward}.

The second layout, Asymmetric Advantages, shares these dynamics, rewards, and horizon. Its two chefs work in separate kitchens that share two central pots, and each kitchen contains its own ingredients, dishes, and serving counter. Because the left chef is closer to a serving counter and the right chef to an onion dispenser, efficient play benefits from complementary roles.

The third task is Level-Based Foraging 2.0.0~\cite{christianos2020shared,christianosLbForaging}, in which two level-1 agents must jointly load three level-2 foods on a fully observed $10\times10$ grid within 50 steps, and the return is normalized to $[0,1]$. Its manual reward is a discounted difference of the distance to the nearest food. The grid size was fixed before the reward search as the smaller of two candidate sizes that allowed measurable learning with room for improvement.

\subsection{Compared Conditions}
Table~\ref{tab:conditions} lists the five compared conditions, which separate the contribution of the incentive module from that of an LLM or a tool-using agent. Besides the task-only and manual rewards, three LLM conditions use the same frozen model, GPT-6.1 Sol, together with the same source code, reward interface, and learner. Their searches start from the manual reward of each task, except that the two agents in Forced Coordination start from native shaping. The Eureka-pattern condition adapts the generate, train, and reflect scheme of Eureka~\cite{ma2024eureka}. For each candidate, a controller assembles a single prompt containing the task, the source code, and the training and reward-component records of earlier candidates, and the model returns one program without tool use. The two agent conditions instead run persistent tool-using sessions~\cite{openaiAgentsSDK} with access to the development archive, numerical analysis, validity checks, and learning probes. They differ only in that the proposed agent receives the incentive module and instructions to investigate incentive failures, so their comparison isolates the contribution of this guidance. In Forced Coordination, the module uses 24 recorded invalid recipe events, and 33 further cases, including valid events, are reserved for assessment.

A defect in how agent sessions handled their turn limit was corrected before the final evaluation, and the affected agent searches were restarted from their initial rewards under unchanged limits.

\subsection{Search and Selection}
All LLM conditions share a maximum reward-search budget of $6B$ per task, which they allocate differently. The Eureka-pattern condition trains two rounds of three candidates for $B$ each, whereas each agent trains two rounds of two candidates for $B$ each and may additionally run up to eight learning probes of $B/4$, with at most two per candidate. Candidates are selected by task return as described in Algorithm~\ref{alg:repair}, using development seeds 0 and 1 and an identical protocol in both Overcooked layouts. The selected programs were then frozen, and new policies were trained for all conditions on previously unused seeds, namely 170--174 in Forced Coordination, 270--274 in Asymmetric Advantages, and 70--74 in Foraging. In Forced Coordination, native shaping was evaluated on the same seeds.

\subsection{Evaluation Metrics}
All metrics are computed from the native task reward. The main outcome is the return of the final policy, averaged over 30 evaluation episodes with sampled actions. Foraging additionally reports the full-completion rate and the completion time, which is set to 50 steps for unsuccessful episodes. Learning speed is summarized by the curve mean, defined as the trapezoidal area under five-episode evaluations at updates 1, 21, \dots, 181, and 200, divided by the length of the observed interval. Behavior is characterized by counts of valid and invalid task events. Episode results are averaged within each policy, and each condition is reported as the mean and sample standard deviation over five training seeds.

\subsection{Additional Experiments}
\label{sec:additional}
Two additional analyses examine the effect of a longer training budget and the role of credit allocation.

\emph{Training budget:} In Forced Coordination, the policies of the manual reward, the Eureka pattern, and both agents were trained further to $2B$ on the same seeds, and the curve mean covers all evaluations up to $2B$. Paired differences between conditions are summarized with 95\% percentile bootstrap intervals over seeds, based on 20,000 resamples~\cite{agarwal2021precipice}.

\emph{Credit allocation:} The role of credit allocation is examined in two ways. First, native shaping and the manual reward contain the same terms and differ only in whether the shaping goes to the acting chef or is shared. Second, a control is applied to the selected Forced Coordination reward, whose placement reward combines an own term $\alpha \ell_i$ and a shared term $\beta\sum_j \ell_j$, where $\ell_i$ denotes the ingredient-placement signal of agent $i$. The control assigns the entire team credit to the acting agent,
\begin{equation}
\rho_i=\alpha \ell_i+\beta\sum_j\ell_j,\qquad \widetilde{\rho}_i=(\alpha+2\beta)\ell_i,
\end{equation}
which leaves the team total unchanged for two agents. With $\alpha=3$ and $\beta=1.5$, a placement by the cook gives 4.5 to the cook and 1.5 to the supplier under the selected reward and 6 to the cook under the control. All other reward terms are retained, and the zero-sum difference $\widetilde{\rho}_i-\rho_i$ is scaled by a common factor that keeps the shaping of each agent within the clipping bound, so that the clipped team shaping is unchanged. The control rule was fixed before any final outcome was observed, and the control was trained on seeds 70--74 with the same learner.

\section{Results}
\label{sec:results}
The results cover the primary comparison, the repair process, longer training, and credit allocation.

\subsection{Primary Comparison}
The agent with the incentive module achieved the highest mean return in all three tasks (Table~\ref{tab:results}). In Forced Coordination, it reached 87.2, compared with 65.6 for the manual reward, 65.1 for the Eureka pattern, and 37.3 for the agent without the module, and it exceeded the latter two in four of five seeds (Table~\ref{tab:longer}).

\begin{table}[!htbp]
\centering
\caption{Final Task Performance\label{tab:results}}
\small
\setlength{\tabcolsep}{3pt}
\renewcommand{\arraystretch}{1.12}
\begin{adjustbox}{max width=\linewidth}
\begin{tabular}{@{}lcccc@{}}
\toprule
 & Forced Coordination & Asymmetric Advantages & \multicolumn{2}{c}{Level-Based Foraging}\\
\cmidrule(lr){2-2}\cmidrule(lr){3-3}\cmidrule(l){4-5}
Condition & Return & Return & Return & Completion (\%)\\
\midrule
Task only & $0.0\pm0.0$ & $0.0\pm0.0$ & $0.704\pm0.316$ & $46.0\pm43.7$\\
Manual & $65.6\pm32.7$ & $194.8\pm113.0$ & $0.664\pm0.215$ & $43.3\pm29.5$\\
Eureka pattern & $65.1\pm15.4$ & $345.3\pm25.8$ & $0.742\pm0.330^{\dagger}$ & $60.0\pm48.7^{\dagger}$\\
Agent without module & $37.3\pm36.2$ & $372.1\pm33.2$ & $0.747\pm0.212^{\dagger}$ & $52.7\pm38.0^{\dagger}$\\
\textbf{Agent with module} & $\mathbf{87.2\pm27.4}$ & $\mathbf{380.4\pm37.6}$ & $\mathbf{0.882\pm0.069}$ & $\mathbf{75.3\pm14.6}$\\
\bottomrule
\end{tabular}
\end{adjustbox}

\vspace{4pt}
\begin{minipage}{\linewidth}\footnotesize
Mean $\pm$ standard deviation over five training seeds. In both Overcooked layouts, the manual reward shares its shaping between the chefs. $^{\dagger}$Best search candidate, reported although it did not pass the confirmation on the second development seed.
\end{minipage}
\end{table}

\begin{table}[!htbp]
\centering
\caption{Paired Comparison in Forced Coordination}
\label{tab:longer}
\scriptsize
\setlength{\tabcolsep}{2pt}

\begin{adjustbox}{max width=\linewidth}
\begin{tabular}{@{}l@{\hspace{10pt}}r@{\,$\pm$\,}lrr@{,\,}l@{\hspace{14pt}}r@{\,$\pm$\,}lrr@{,\,}l@{\hspace{14pt}}r@{\,$\pm$\,}lrr@{,\,}l@{}}
\toprule
 & \multicolumn{5}{c}{Return at $B$} & \multicolumn{5}{c}{Return at $2B$} & \multicolumn{5}{c}{Curve mean}\\
\cmidrule(r{7pt}){2-6}\cmidrule(lr{7pt}){7-11}\cmidrule(l){12-16}
Reward & \multicolumn{2}{c}{Mean $\pm$ SD} & Gain & \multicolumn{2}{c}{95\% CI} & \multicolumn{2}{c}{Mean $\pm$ SD} & Gain & \multicolumn{2}{c}{95\% CI} & \multicolumn{2}{c}{Mean $\pm$ SD} & Gain & \multicolumn{2}{c}{95\% CI}\\
\midrule
Manual & 65.6 & 32.7 & $+21.6$ & [$-8.8$ & 50.3] & 152.5 & 30.2 & $+15.5$ & [$-9.6$ & 44.3] & 73.3 & 18.1 & $+15.2$ & [0.1 & 34.4]\\
Eureka pattern & 65.1 & 15.4 & $+22.1$ & [2.5 & 41.7] & 163.1 & 23.1 & $+4.9$ & [$-13.7$ & 28.0] & 75.9 & 4.5 & $+12.6$ & [4.8 & 18.5]\\
Agent without module & 37.3 & 36.2 & $+49.9$ & [5.9 & 93.9] & 101.9 & 94.9 & $+66.1$ & [$-0.4$ & 132.7] & 42.7 & 40.8 & $+45.8$ & [12.0 & 79.5]\\
\midrule
\textbf{Agent with module} & \textbf{87.2} & \textbf{27.4} & \multicolumn{3}{c}{--} & \textbf{168.0} & \textbf{12.8} & \multicolumn{3}{c}{--} & \textbf{88.5} & \textbf{9.9} & \multicolumn{3}{c}{--}\\
\bottomrule
\end{tabular}
\end{adjustbox}
\vspace{3pt}
\begin{minipage}{\linewidth}
\footnotesize
Mean $\pm$ standard deviation over five training seeds. Gain is the paired difference between the agent with module and the reward in each row, with a 95\% percentile bootstrap confidence interval. The Eureka search started from the manual reward and the agent searches from native shaping.
\end{minipage}
\end{table}

\begin{figure}[!htbp]
\centering
\includegraphics[width=\linewidth,height=0.88\textheight,keepaspectratio]{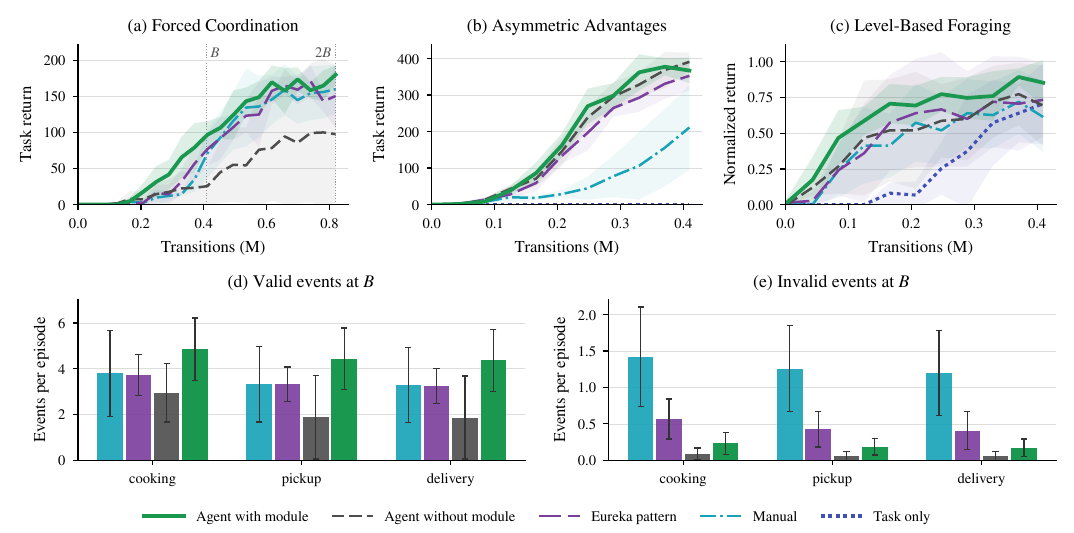}
\caption{Learning curves and behavior. (a)--(c) Task return during training, with Forced Coordination trained to $2B$. (d) Valid and (e) invalid recipe events per episode in Forced Coordination at $B$. Curves and bars show the mean and one standard deviation over five seeds.\label{fig:results}}
\end{figure}

Behavior explains this advantage (Fig.~\ref{fig:results}d,e). Relative to the Eureka pattern, the selected reward produced about one more valid cooking start, soup pickup, and delivery per episode, with fewer invalid starts and pickups, and relative to the manual reward it reduced invalid deliveries from 1.20 to 0.17 per episode. In Foraging, the agent with the module reached a normalized return of 0.882 and a full-completion rate of 75.3\%, compared with 0.747 and 52.7\% for the agent without the module, 0.742 and 60.0\% for the Eureka pattern, and 0.664 and 43.3\% for the manual reward (Fig.~\ref{fig:results}c). Its results also varied much less across seeds. The manual reward learned faster than the task reward alone, with nearly twice its curve mean, but finished slightly lower, 0.664 compared with 0.704, a difference well within the variation across seeds. Hand-designed shaping thus accelerated learning without improving the final outcome, whereas the repaired reward improved both.

In Asymmetric Advantages, where all searches started from the manual reward, both agents clearly exceeded the manual reward and the Eureka pattern, which reached 194.8 and 345.3 (Fig.~\ref{fig:results}b). The agent with the module again obtained the highest return, 380.4, and the highest curve mean. Its advantage over the agent without the module, which reached 372.1, was smaller in this layout, because that agent also added terms for valid and invalid cooking starts and thereby addressed the same recipe problem.

\subsection{Diagnosis and Repair}
The incentive module explains why native shaping failed in Forced Coordination and how the selected reward corrected it. Native shaping pays $+5$ for picking up a cooked soup even when the soup cannot be delivered, and it is indifferent to starting an incomplete recipe (first row of Table~\ref{tab:repair}). Policies trained with it learned to exploit this bonus. In 150 evaluation episodes on seeds 70--74, all 407 delivered soups were invalid, and 386 of them contained a single onion.

Table~\ref{tab:repair} traces how the agent used this evidence. In the first round, penalizing both invalid events removed the bonus but produced no return, which shows that correcting local preferences is not sufficient on its own. Adding shared credit for useful ingredient placements, together with a small cost for holding a dish too early, then enabled the chefs to deliver, although the pickup gap remained slightly positive. In the second round, both candidates revised this program and made the pickup gap negative, and the one that also credited the supplier when a valid soup is collected achieved the highest return and was selected. On separate assessment cases, the selected reward also gives $+4$ for valid cooking and keeps $+5$ for valid pickups.

\begin{table}[!htbp]
\centering
\caption{Step-by-Step Reward Repair in Forced Coordination\label{tab:repair}}
\small
\setlength{\tabcolsep}{3.4pt}
\renewcommand{\arraystretch}{1.12}
\begin{adjustbox}{max width=\linewidth}
\begin{tabular}{@{}lrrrr@{}}
\toprule
 & \multicolumn{2}{c}{Gap on invalid events} & \multicolumn{2}{c}{Development run}\\
\cmidrule(lr){2-3}\cmidrule(l){4-5}
Reward & Cooking & Pickup & Return & Invalid\\
\midrule
Native shaping & $0.00$ & $+5.00$ & 2.0 & 2.60\\
\multicolumn{5}{@{}l}{\emph{Round 1}}\\
\quad Penalize invalid events & $-1.00$ & $-0.50$ & 0.0 & 0.00\\
\quad Add shared placement credit & $-0.50$ & $+0.01$ & 110.7 & 0.13\\
\multicolumn{5}{@{}l}{\emph{Round 2}}\\
\quad Correct pickup gap & $-0.50$ & $-0.01$ & 111.3 & 0.07\\
\quad \textbf{Correct gap, credit supplier} & $\mathbf{-0.50}$ & $\mathbf{-0.01}$ & \textbf{128.7} & \textbf{0.27}\\
\bottomrule
\end{tabular}
\end{adjustbox}

\vspace{4pt}
\begin{minipage}{\linewidth}\footnotesize
Gap is the reward for the invalid event minus that for staying in place, so negative values discourage it. Development runs use seed 0, and bold marks the selected reward.
\end{minipage}
\end{table}

In Foraging, the module showed that the manual reward credited an agent for approaching its own nearest food even when this moved the team away from the common target, with a mean of $+0.056$ on such transitions. The selected reward assigns $-0.010$ to these transitions and raises the credit for common-target progress from $+0.031$ to $+0.086$.

\subsection{Longer Training}
With twice the training budget, all Forced Coordination rewards improved and the alternatives moved closer, but the selected reward still attained the highest final return, 168.0, with the smallest variation across seeds (Table~\ref{tab:longer} and Fig.~\ref{fig:results}a). It also achieved the highest curve mean, 88.5, compared with 75.9 for the Eureka pattern and 73.3 for the manual reward, which reflects faster learning. The agent without the module reached 101.9, because two of its runs produced almost no return.

\subsection{Credit Allocation}

Table~\ref{tab:credit} isolates who receives the intermediate credit. In each row, the reward terms and the team total are unchanged, and only the recipient differs. With shared credit, cooperation was learned, whereas with credit to the acting chef only, the return fell to 0.4 in both cases, with a decrease in every seed. For the selected reward, which was trained on seeds 70--74 for this test, viable ingredient placements fell from 13.8 to 2.1 per episode, and a replay of all 120,000 final evaluation transitions confirmed that the team shaping was unchanged. Because only the supplier can provide ingredients in this layout, it stops supplying them once it receives no share of the credit.

\begin{table}[h!]
\centering
\caption{Effect of Credit Allocation in Forced Coordination\label{tab:credit}}
\small
\setlength{\tabcolsep}{3pt}
\renewcommand{\arraystretch}{1.12}
\begin{adjustbox}{max width=\linewidth}
\begin{tabular}{@{}lccc@{}}
\toprule
 & \multicolumn{2}{c}{Final return} & \\
\cmidrule(lr){2-3}
Reward & Shared credit & Acting chef only & Drop\\
\midrule
Manual reward & $65.6\pm32.7$ & $0.4\pm0.6$ & $65.2$ \ $[40.9,\ 91.6]$\\
Selected reward & $68.1\pm20.1$ & $0.4\pm0.9$ & $67.7$ \ $[52.5,\ 83.6]$\\
\bottomrule
\end{tabular}
\end{adjustbox}
\vspace{4pt}
\begin{minipage}{\linewidth}\footnotesize
Mean $\pm$ standard deviation over five training seeds. Each row changes only which chef receives the intermediate credit, and the drop is a paired difference with a 95\% bootstrap interval.
\end{minipage}
\end{table}

\section{Discussion}
\label{sec:discussion}
The results show that explicit incentive measurements improve LLM-based reward design. With the same model, tools, and budget, the agent with the module obtained the highest return in all three tasks. Its advantage was largest in Forced Coordination and Foraging, where the explicit measurements led to targeted repairs that learned more reliably than the rewards found without them.

The repair trace and the credit comparisons show how these gains arise. Correcting local preferences alone did not produce learning, whereas restoring the supplier's share of placement credit did, and removing that share eliminated cooperation. The contrast between native shaping and the manual reward supports the same conclusion. Under agent-specific rewards, who receives intermediate credit matters as much as what is rewarded, and the proposed agent found this allocation together with the recipe repair without manual intervention.

This study also has limitations that point to directions for future work. Each reward search was run once per condition with five training seeds, so repeated searches and additional seeds would strengthen the statistical comparisons. The diagnostics were designed for each task, and all experiments involved two agents and a single LLM, which makes automatically constructed diagnostics, larger teams, and other language models natural extensions. Finally, the effect of credit allocation applies to learners that keep agent-specific rewards, because a learner that optimizes a single team reward does not distinguish between allocations.

\section{Conclusions}
\label{sec:conclusion}
This paper introduced incentive-aware reward repair, in which fixed measurements of action preference and cooperative credit guide an LLM in revising executable reward programs for cooperative MARL. Across two Overcooked layouts and Level-Based Foraging, the repaired rewards achieved the highest task return, improved on an otherwise identical LLM agent without these measurements, and learned faster and more reliably than the alternatives. Comparisons of shared and individual credit further showed that, with the team total fixed, the agent that receives intermediate credit can determine whether cooperation emerges. Cooperative reward design should therefore address both what is rewarded and who receives the credit, and incentive-aware diagnostics offer a practical way to achieve this with LLMs.

\section*{Acknowledgment}
This work was supported by the Scientific and Technological Research Council of T\"urkiye (T\"UB\.ITAK) under Project No. 124E305.

\bibliographystyle{unsrt}
\bibliography{references}

@inproceedings{ma2024eureka,
  author    = {Yecheng Jason Ma and William Liang and Guanzhi Wang and De-An Huang and Osbert Bastani and Dinesh Jayaraman and Yuke Zhu and Linxi Fan and Anima Anandkumar},
  title     = {{Eureka}: Human-level reward design via coding large language models},
  booktitle = {Proc. Int. Conf. Learn. Represent. (ICLR)},
  year      = {2024}
}

@inproceedings{xie2024text2reward,
  author    = {Tianbao Xie and Siheng Zhao and Chen Henry Wu and Yitao Liu and Qian Luo and Victor Zhong and Yanchao Yang and Tao Yu},
  title     = {{Text2Reward}: Reward shaping with language models for reinforcement learning},
  booktitle = {Proc. Int. Conf. Learn. Represent. (ICLR)},
  year      = {2024}
}

@inproceedings{ng1999shaping,
  author    = {Andrew Y. Ng and Daishi Harada and Stuart J. Russell},
  title     = {Policy invariance under reward transformations: Theory and application to reward shaping},
  booktitle = {Proc. 16th Int. Conf. Mach. Learn. (ICML)},
  year      = {1999},
  pages     = {278--287}
}

@inproceedings{foerster2018counterfactual,
  author    = {Jakob Foerster and Gregory Farquhar and Triantafyllos Afouras and Nantas Nardelli and Shimon Whiteson},
  title     = {Counterfactual multi-agent policy gradients},
  booktitle = {Proc. AAAI Conf. Artif. Intell.},
  volume    = {32},
  number    = {1},
  year      = {2018},
  doi       = {10.1609/aaai.v32i1.11794}
}

@inproceedings{yu2022surprising,
  author    = {Chao Yu and Akash Velu and Eugene Vinitsky and Jiaxuan Gao and Yu Wang and Alexandre Bayen and Yi Wu},
  title     = {The surprising effectiveness of {PPO} in cooperative multi-agent games},
  booktitle = {Proc. Adv. Neural Inf. Process. Syst. (NeurIPS)},
  volume    = {35},
  pages     = {24611--24624},
  year      = {2022}
}

@inproceedings{carroll2019overcooked,
  author    = {Micah Carroll and Rohin Shah and Mark K. Ho and Tom Griffiths and Sanjit A. Seshia and Pieter Abbeel and Anca D. Dragan},
  title     = {On the utility of learning about humans for human-{AI} coordination},
  booktitle = {Proc. Adv. Neural Inf. Process. Syst. (NeurIPS)},
  volume    = {32},
  pages     = {5175--5186},
  year      = {2019}
}

@misc{christianosLbForaging,
 author={Filippos Christianos},
 title={Level-Based Foraging},
 howpublished={\url{https://github.com/semitable/lb-foraging}},
 note={Version 2.0.0}
}

@inproceedings{christianos2020shared,
  author    = {Filippos Christianos and Lukas Sch{\"a}fer and Stefano V. Albrecht},
  title     = {Shared experience actor-critic for multi-agent reinforcement learning},
  booktitle = {Proc. Adv. Neural Inf. Process. Syst. (NeurIPS)},
  volume    = {33},
  year      = {2020}
}

@misc{openaiAgentsSDK,
 author={{OpenAI}},
 title={{OpenAI Agents SDK}},
 howpublished={\url{https://openai.github.io/openai-agents-python/}},
 note={Version 0.8.4}
}

@misc{overcookedRewardWrapper,
 author={{HumanCompatibleAI}},
 title={{Overcooked-AI}: Event-Reward Multi-Agent Wrapper},
 howpublished={\url{https://github.com/HumanCompatibleAI/overcooked_ai/blob/master/src/human_aware_rl/rllib/rllib.py}},
 note={Accessed November 2026}
}

@inproceedings{agarwal2021precipice,
  author    = {Rishabh Agarwal and Max Schwarzer and Pablo Samuel Castro and Aaron Courville and Marc G. Bellemare},
  title     = {Deep reinforcement learning at the edge of the statistical precipice},
  booktitle = {Proc. Adv. Neural Inf. Process. Syst. (NeurIPS)},
  volume    = {34},
  year      = {2021}
}

@article{gronauer2022survey,
  author  = {Sven Gronauer and Klaus Diepold},
  title   = {Multi-agent deep reinforcement learning: A survey},
  journal = {Artif. Intell. Rev.},
  volume  = {55},
  number  = {2},
  pages   = {895--943},
  year    = {2022},
  doi     = {10.1007/s10462-021-09996-w}
}

@article{wong2023deep,
  author  = {Annie Wong and Thomas B{\"a}ck and Anna V. Kononova and Aske Plaat},
  title   = {Deep multiagent reinforcement learning: Challenges and directions},
  journal = {Artif. Intell. Rev.},
  volume  = {56},
  pages   = {5023--5056},
  year    = {2023},
  doi     = {10.1007/s10462-022-10299-x}
}

@book{sutton2018reinforcement,
  author    = {Richard S. Sutton and Andrew G. Barto},
  title     = {Reinforcement Learning: An Introduction},
  edition   = {2nd},
  publisher = {MIT Press},
  address   = {Cambridge, MA, USA},
  year      = {2018}
}

@inproceedings{devlin2011theory,
  author    = {Sam Devlin and Daniel Kudenko},
  title     = {Theoretical considerations of potential-based reward shaping for multi-agent systems},
  booktitle = {Proc. 10th Int. Conf. Auton. Agents Multiagent Syst. (AAMAS)},
  year      = {2011},
  pages     = {225--232}
}

@inproceedings{hadfieldmenell2017ird,
  author    = {Dylan Hadfield-Menell and Smitha Milli and Pieter Abbeel and Stuart Russell and Anca D. Dragan},
  title     = {Inverse reward design},
  booktitle = {Proc. Adv. Neural Inf. Process. Syst. (NeurIPS)},
  volume    = {30},
  year      = {2017}
}

@article{wolpert2002coin,
  author  = {David H. Wolpert and Kagan Tumer},
  title   = {Collective intelligence, data routing and {Braess'} paradox},
  journal = {J. Artif. Intell. Res.},
  volume  = {16},
  pages   = {359--387},
  year    = {2002},
  doi     = {10.1613/jair.995}
}

@inproceedings{devlin2014pbdR,
  author    = {Sam Devlin and Logan Yliniemi and Daniel Kudenko and Kagan Tumer},
  title     = {Potential-based difference rewards for multiagent reinforcement learning},
  booktitle = {Proc. 13th Int. Conf. Auton. Agents Multiagent Syst. (AAMAS)},
  year      = {2014},
  pages     = {165--172}
}

@inproceedings{li2025rstar,
  author    = {Pengyi Li and others},
  title     = {{R}*: Efficient reward design via reward structure evolution and parameter alignment optimization with large language models},
  booktitle = {Proc. 42nd Int. Conf. Mach. Learn. (ICML)},
  series    = {Proc. Mach. Learn. Res.},
  volume    = {267},
  pages     = {34509--34527},
  year      = {2025}
}

@inproceedings{gao2025rfagent,
  author    = {N. Gao and X. Zhang and X. Jiang and M. You and M. Zhang and Y. Deng},
  title     = {{RF-Agent}: Automated reward function design via language agent tree search},
  booktitle = {Proc. Adv. Neural Inf. Process. Syst. (NeurIPS)},
  volume    = {38},
  year      = {2025}
}

@article{zhu2025lamarl,
  author  = {G. Zhu and R. Zhou and W. Ji and S. Zhao},
  title   = {{LAMARL}: {LLM}-aided multi-agent reinforcement learning for cooperative policy generation},
  journal = {IEEE Robot. Autom. Lett.},
  volume  = {10},
  number  = {7},
  pages   = {7476--7483},
  month   = jul,
  year    = {2025},
  doi     = {10.1109/LRA.2025.3577527}
}

@inproceedings{qu2025latent,
  author    = {Y. Qu and others},
  title     = {Latent reward: {LLM}-empowered credit assignment in episodic reinforcement learning},
  booktitle = {Proc. AAAI Conf. Artif. Intell.},
  volume    = {39},
  number    = {19},
  pages     = {20095--20103},
  year      = {2025},
  doi       = {10.1609/aaai.v39i19.34213}
}

@article{wei2026platoon,
  author  = {D. Wei and P. Yi and J. Lei and Y. Hong and H. Dong and Y. Du},
  title   = {An automated reinforcement learning reward design framework with large language model for cooperative platoon coordination},
  journal = {IEEE Trans. Intell. Transp. Syst.},
  volume  = {27},
  number  = {7},
  pages   = {8252--8269},
  month   = jul,
  year    = {2026},
  doi     = {10.1109/TITS.2026.3664858}
}

@inproceedings{papoudakis2021benchmarking,
  author    = {Georgios Papoudakis and Filippos Christianos and Lukas Sch{\"a}fer and Stefano V. Albrecht},
  title     = {Benchmarking multi-agent deep reinforcement learning algorithms in cooperative tasks},
  booktitle = {Proc. Neural Inf. Process. Syst. Track Datasets Benchmarks},
  volume    = {1},
  year      = {2021}
}

@misc{overcookedAI,
  author       = {{HumanCompatibleAI}},
  title        = {{Overcooked-AI}},
  howpublished = {\url{https://github.com/HumanCompatibleAI/overcooked_ai}},
  note         = {Version 1.1.0}
}

@inproceedings{tan1993independent,
  author    = {Ming Tan},
  title     = {Multi-agent reinforcement learning: Independent vs. cooperative agents},
  booktitle = {Proc. 10th Int. Conf. Mach. Learn. (ICML)},
  year      = {1993},
  pages     = {330--337}
}

@misc{li2026arbo,
  author       = {M. Li and R. Tan and R. Wang},
  title        = {Beyond scripted search: Sample-efficient reward discovery via agentic black-box optimization},
  howpublished = {arXiv:2609.32394},
  year         = {2026}
}

@inproceedings{sunehag2018value,
  author    = {Peter Sunehag and Guy Lever and Audrunas Gruslys and Wojciech Marian Czarnecki and Vinicius Zambaldi and Max Jaderberg and Marc Lanctot and Nicolas Sonnerat and Joel Z. Leibo and Karl Tuyls and Thore Graepel},
  title     = {Value-decomposition networks for cooperative multi-agent learning based on team reward},
  booktitle = {Proc. 17th Int. Conf. Auton. Agents Multiagent Syst. (AAMAS)},
  year      = {2018},
  pages     = {2085--2087}
}

@inproceedings{harutyunyan2015pba,
  author    = {Anna Harutyunyan and Sam Devlin and Peter Vrancx and Ann Now{\'e}},
  title     = {Expressing arbitrary reward functions as potential-based advice},
  booktitle = {Proc. AAAI Conf. Artif. Intell.},
  volume    = {29},
  pages     = {2652--2658},
  year      = {2015}
}

@article{rashid2020qmix,
  author  = {Tabish Rashid and Mikayel Samvelyan and Christian Schroeder de Witt and Gregory Farquhar and Jakob Foerster and Shimon Whiteson},
  title   = {Monotonic value function factorisation for deep multi-agent reinforcement learning},
  journal = {J. Mach. Learn. Res.},
  volume  = {21},
  number  = {178},
  pages   = {1--51},
  year    = {2020}
}

\end{document}